# A Multi-Purpose Scenario-based Simulator
# for Smart House Environments

Zahra Forootan Jahromi and Amir Rajabzadeh*
Department of Computer Engineering
Razi University
Kermanshah, Iran
zahra.forootan@gmail.com, rajabzadeh@razi.ac.ir

Ali Reza Manashty
Department of IT and Computer Engineering
Shahrood University of Technology
Shahrood, Iran
a.r.manashty@gmail.com

*Abstract:* **Developing smart house systems has been a great challenge for researchers and engineers in this area because of the high cost of implementation and evaluation process of these systems, while being very time consuming. Testing a designed smart house before actually building it is considered as an obstacle towards an efficient smart house project. This is because of the variety of sensors, home appliances and devices available for a real smart environment. In this paper, we present the design and implementation of a multi-purpose smart house simulation system for designing and simulating all aspects of a smart house environment. This simulator provides the ability to design the house plan and different virtual sensors and appliances in a two dimensional model of the virtual house environment. This simulator can connect to any external smart house remote controlling system, providing evaluation capabilities to their system much easier than before. It also supports detailed adding of new emerging sensors and devices to help maintain its compatibility with future simulation needs. Scenarios can also be defined for testing various possible combinations of device states; so different criteria and variables can be simply evaluated without the need of experimenting on a real environment.**

## I. INTRODUCTION

As new technologies are emerging, people are more eager to apply these technologies to their house in order to be more and more comfortable and secure. Smart houses, as a state-of-the-art technology in two last decades, are becoming the most exciting and useful tools in our daily lives, which has brought a higher comfort and security level into our life.

The terms smart homes and intelligent homes have been used for more than a decade to introduce the concept of smart devices and equipment in the house. According to the Smart Homes Association the best definition of the smart home technologies is "The integration of technology and services through home networking for a better quality of living".

Smart home is not only an interesting topic, but also a burgeoning industry as well as entering to a broad audience home gradually [1]. Most programmers have to design smart home systems case by case and spend a lot of time managing them [2]. Many others have already presented how to cut down the building costs by using smart home simulators or high level programming languages [3].

Smart houses could be divided into two main categories:

- Programmable houses – are those scenario-based systems programmed to perform an action triggered by a condition on a sensor output.

- Intelligent houses – are those that possess some kind of intelligence without the need of precise manual design of the procedures.

### A. Programmable Houses

Programmable houses will be those that have reactions based only on simple sensor inputs, and possess no built-in intelligence. Such a house for a predefined input has a programmed set of actions to perform.

Examples of such actions might be light bulbs operated by movement sensors, or selection of one of the predefined lighting settings by a button on a remote controller.

Actually, many of currently manufactured and sold smart house systems belong to this group.

The biggest problem with this type of houses is that they have to be reprogrammed when some of the features change. That presents a problem for many people and requires calling a technician to get the job done.

Hence increasing tension to develop some smart home solution that is based on artificial intelligence will adapt its operation to changing user behavior. That tension leads to development of the houses that belong to the second category. This paper, though, supports the first group of smart houses described earlier.

### B. Inteligent Houses

They represent the state-of-the-art technology. Those types of installations are driven by artificial intelligence, and instead of having to be programmed they are able to learn basing on observation of inhabitants behavior over a period of time.

*Corresponding Author

One of the first successful implementations was well known Adaptive House developed by M. Mozer at University of Colorado back in 1998. Some other examples that belong to the group of intelligent houses are:

• Georgia Tech Aware Home
• AIRE spaces at MIT
• Interactive Workspaces Project at Stanford
• Gaia project at UIUC
• MavHome project at UTA

The smart house consists of a large and wide ranging set of many services, applications, equipment, networks and systems that act together in delivering the "intelligent" or "connected" home in order to maintain security and control, communications, leisure and comfort, environmental integration and accessibility. These components are represented by many actors that interact and work together to provide interactive systems that benefit the home based user in the smart house. Because of this wide ranging variability of the entities in the smart house, there is a very high level of potential complexity in finding the optimal solution for each different smart house.

For researchers and engineers, it is difficult to work in the real smart home since home appliances are very expensive.

In this paper we present the designing and implementation of a comprehensive smart house simulator to reduce these complexities of implementation a smart house and also find the best solution of making a home or a building smart. Our simulator is completely object based, because we have considered no limitation in different process of simulation.

## II. RELATED WORKS

There have been lots of works on this research area including the big corporations and research groups. As a result, various ubiquitous computing simulators such as the Ubiquitous Wireless Infrastructure Simulation Environment (Ubiwise) and TATUS and Context Aware Simulation Toolkit (CAST) have been proposed. The Ubiwise Simulator is used to test computation and communication devices. It has three dimensional (3D) models that form a physical environment viewed by users on a desktop computer through two windows [4, 5]. This simulator focuses on device testing, e.g., in aggregating device functions and exploring the integration of handheld devices and Internet service. Thus, this simulator does not consider an adaptive environment. TATUS is built using the Half Life game engine. Therefore, it looks like an assembled simulation game. It constructs a 3D virtual environment, e.g., a meeting scenario. Using this simulator, a user commands a virtual character to perform tasks, such as to sit down. This simulator does not consider device simulation [6]. CAST is a simulator for the test home domain. This simulator uses scenario based approach. It has been proposed as a prototype using Macromedia's Flash MX 2004 [7]. However, using Flash MX [8] does not support users to freely control their environment. Joon Seok Park et al. proposed the design structure for smart home simulator regardless of environment factor as well as interaction aspect [9].

## III. PROPOSED SMART HOUSE SIMULATOR

There are many simulators in different scope of science and the main purpose of implementing and developing them is demonstrating a virtual model of real subject as well, in order to decrease the problems and difficulties emerge in the way of implementing and evaluating the proposed project in reality.

Indeed researchers use simulators to decrease costs and consumed time for testing and evaluating their ideas on developing and evaluating a project. So the principle duty of a simulator is simulating a virtual model of reality that must be close to its actual model in the real world

In this paper, we present the designing and implementation of smart house simulator for developing and evaluating smart house projects to decrease the obstacles in the way of such projects, mostly cost and time. Due to some difficulties such as providing the necessary real sensors and home appliances to analyze the real home environment, couldn't advance any further than their design level.

This simulator can be used as a substitution for the corresponding real smart environment. Every kind of state-of-the-art sensors and home appliances can be used in the proposed simulator. All the necessary requirements for making a house smart are provided in the simulator.

In the following sections we explain the designing and implementation level of the project and then discuss about the main features of the proposed simulator.

All the principle futures and main capabilities are considered in the designing level, which distinct the proposed simulator from other similar systems.

Some of the most important characteristics of the simulator are describing in the following sections. These principle features of the proposed system are illustrated in Fig.1.

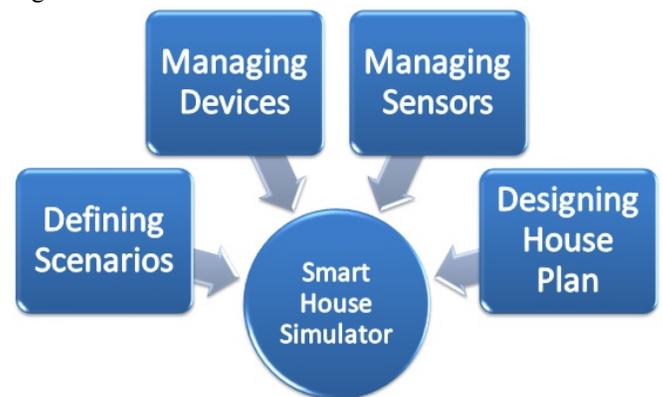

Figure 1: Principle features of the proposed Smart House Simulator

### A. Top view plan of the specified house

The simulator should have the capability of demonstrating the plane of the desired house plan in order to be able to simulate a more real virtual model of the house (Fig. 2). The possibility of drawing the house plane is provided in this simulator, so the user can define all boundaries of the house such as different rooms, doors, windows and etc. The user also can load an image of top view plan of a house as the house plan.

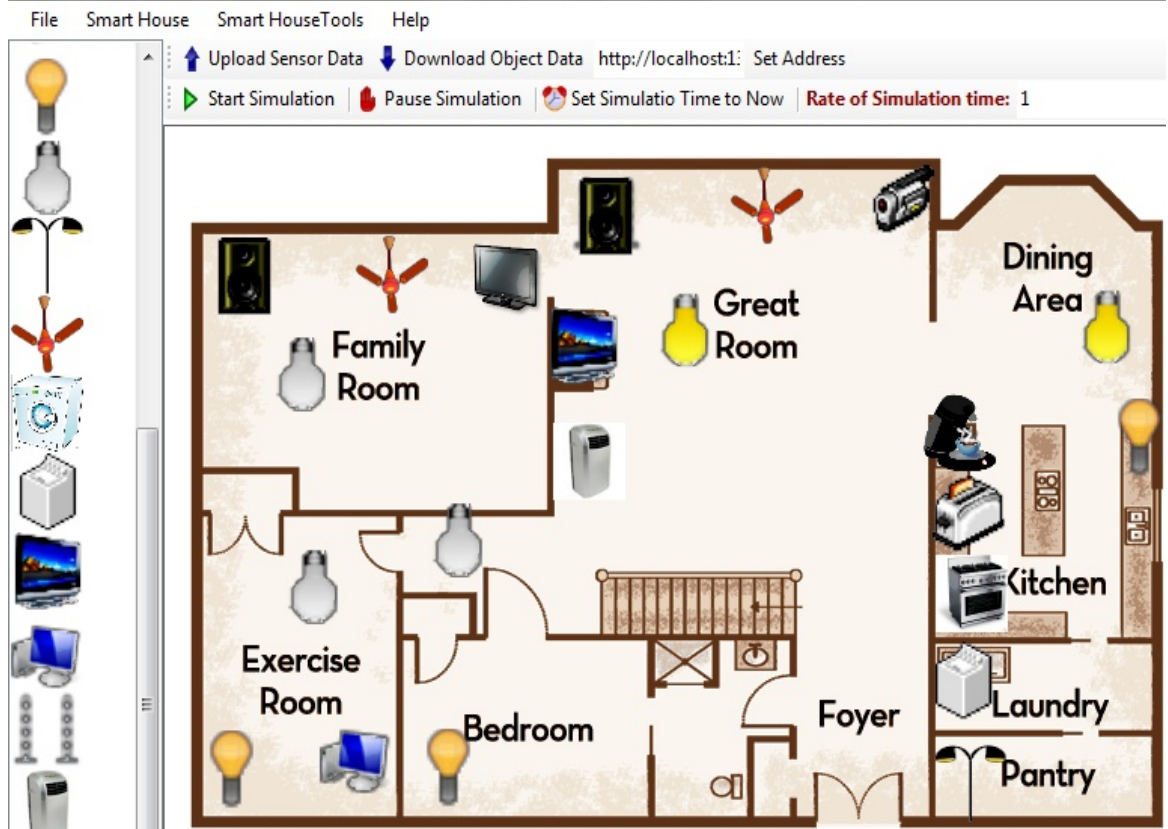

Figure 2: Simulator environment containing top view plan of the house and virtual simulated devices

After designing the house plan user should place each home appliance in their positions as they placed is in real house so user can distinguish them easily for crating different tasks in different objects. User can design the most real model of the real house by using this capability of the proposed simulator.

### B. Supporting all kinds of sensors and home appliances

Using different types of sensors and actuators for getting and setting status of each device is an inseparable part of every smart house projects. Many of these sensors are too expensive and some of them have various kinds with different futures of a certain type.

As technology is improving so fast, it's obvious that every day a new kind of sensors, actuators and home appliances will emerge, so the ability of supporting any kind of sensors and actuators is an important future for a smart house simulator.

This simulator provides the possibility of creating a virtual model of any kind of cutting-edge sensors and devices with defining all of their details like the kind of data that each sensor can sense (Fig. 3).

As it is shown in this figure, first user should enter a name for the considered sensor and then select the data format of it. Data format is the format of the considered sensor that each sensor uses it for demonstrating the status of environment. We have considered 3 data format for sensors contains "Numeral", "Point" and "Multi States", because almost all the sensors data format can be in one of these kinds of data. For example light sensor data format is multi state and it means this sensor use for example two state of "On" and "Off" for showing the light status of environments.

The data format of each sensor can be defined via this form, so that every kind of sensors will all details can be simulated and have a very close model of each sensor in order to have an optimal simulation of smart houses.

For example a light sensor demonstrate the level of light by describing it in 3 level of light, dim and dark; but a temperature sensor show the temperature of an environment in range of numbers or a temperature sensor demonstrates the temperature status of the environments in a range of number, so user should choose the numeral data format for this kind of sensor.

So there is no limitation for using any kind of off the shelf equipment for simulating a virtual smart house environment as well and user can add any number and kind of sensors and home appliances in the simulator. Fig. 4 illustrates the Windows Form which handles adding any virtual model of devices to the simulator. Then user should assign related sensors to each device and choose the devices icon to be shown in designed house environment.

Figure 3: Add sensor form. This form enables the user to create any kind of sensors for use in the devices.

Figure 4: Adding house devices form

### C. Connecting to house remote controlling systems

There are a number of houses remote controlling projects which controlling the devices of a smart house remotely via web or mobile messaging systems.

The proposed simulator can connect to the server of these systems through a network or web. Users can observe the designing simulated house by using internet or via a network and define a task to be done and then send it as a command to the house remote controlling server. The simulator is always checking the server and applies the commands on proper devices.

After each tasks done, the simulator send the updated status of each device to the server. The format of updated status for sending to the server should be in a certain format as Fig. 5 illustrates. This feature enables the use of this simulator as a good substitution for smart houses testing facilities.

| Object_ID | Sensor_ID | Sensor_Value | TimeStamp |
|---|---|---|---|

Figure 5: The fields in the packet used to send data to smart house remote controlling server

### D. Planning scenarios

Scenarios make it easy for people saving the list of actions for further use, in addition to design multiple actions to be done in a single scenario. Later the scenarios can be enabled / disabled in the scenarios list or be used in another scenario too. Cheng, Wang and Chen proposed a reasoning system for smart houses that is also scenario based [10].

One of the capabilities of this simulator, which distinct it from other smart house simulators is the ability of creating scenarios. Each scenario consists of some scheduled tasks and each scheduled task defines a particular action for executing on a special device. As it is shown in figure 5, each scenario has a name and first executing time and date. It means for the first time that a scenario will be executed, user should define the date and time of executing it. A repeating time is considered to repeat the scenario automatically after the first time it executed, and there is no need that user each time Enable the scenario.

So by using a scenario, a set of tasks will execute continuously. As it is shown in the Fig. 6, a delay time is considered for each task which was selected to be added in a scenario. This means as soon as a scenario executes, a set of selected tasks will be run after its defined time passed from the previous executed task.

User can define each schedule and set a combination of specified schedules as a scenario. Scenarios are used for testing various possible combinations of device states; so different criteria and variables can be simply evaluated, without the need of experimenting on a real environment. The scenarios are designed to set a number of tasks all in one place for further and easier use [11].

## IV. EVALUATION

To evaluate the proposed Smart House Simulator we considered a set of scheduled tasks which created by user via the proposed software and then execute them on defined device at the defined time.

In order to create a test plan for simulating via proposed simulator, first user should design a house plane and defines different home appliances and sensors and assign related sensors to each appliance. Then via the "define scheduled task" some tasks should be created on desired objects which are used in the house.

Each task executes on a special device at a specific time and date and set the device's sensor to defined data. Also a scenario can be created via proposed smart house simulator as it described in planning scenarios section.
Scenarios are a set of these tasks which user has create them via "scheduled a task" form without considering their Time and date.

Figure 6: Define Scenario form. A set of scheduled task in the "Scheduled Tasks" checked box, should be selected for defining the scenario. These scheduled are created by user via "Define Schedule" form.

To check the updated status of devices trough the simulator, a solution considered that can show status of each device every time the user get the status of them. As it is demonstrated in Fig. 7, by clicking on a specific device and choose "Get Status" option, the updated status of the selective device shown in a message box. So it can be realized if each task has executed correctly or not. Testing a defined scenario is the same also. We only should get the status of all devices which are defined in the scenario as scheduled tasks.

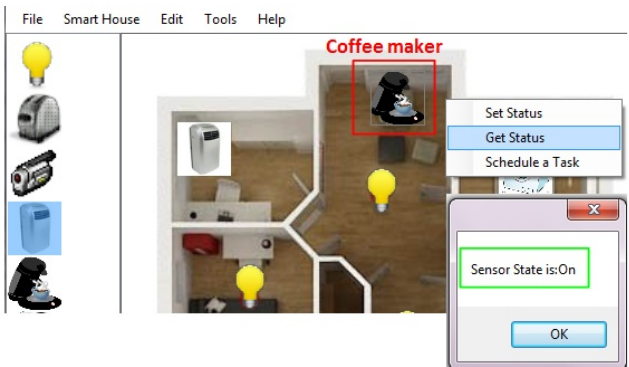

Figure 7: checking the status of selected device

## V.   CONCOLUSIONS

In  this paper we presented design and implementation of smart house simulator, which has considered all aspect of smart environment and there is no limitation for using virtual state-of-the-art sensors and appliances for simulation.

The proposed simulator have some characteristics which distinct it from other smart house simulators. The ability of designing the plan of desired house plan or load top view plan of the house as it's plan and then placed appliances in their real place, makes this simulator to simulate a house very close to the real mod.

Scenarios make it easy for people saving the list of actions for further use, in addition to design multiple actions to be done in a single scenario.

The ability of connecting to smart house remote controlling systems is a good choice for researchers in this area to evaluate their projects. Because this feature of propose smart house simulator need a special format for interpreting commands, we considered a unique format for commands which received from house remote controlling systems.

## VI.   FUTURE WORKS

In the future works, we plan to add the ability to define some variables such as energy consumption, computational complexity and etc. so that we can observe the changes in these variables as the simulation continues. This can be a great leap forward in minimization plans that are about to be applied on real smart houses but the real outcome of plan is hard to estimate as a linear or even nonlinear equation. The simulator can run the house scenarios and calculate real-time values of the variables so we can have a better estimate of the minimization plan if applied to a similar real house.

## AUTHORS PROFILE

Amir Rajabzadeh received the B.S. degree in telecommunication engineering from Tehran University, Iran, in 1990 and received the M.S. and Ph.D. degrees in computer engineering from Sharif University of Technology, Iran, in 1999 and 2005, respectively. He is currently an assistant professor of Computer Engineering at Razi University, Kermanshah-Iran. He was the Head of Computer Engineering Department (2005–2008) and the Education and Research Director of Engineering Faculty (2008-2010) at Razi University.

Zahra Forootan Jahromi is a senior B.S. student in Software Computer Engineering at Razi University, Kermanshah. She is now researching in smart environments specially on simulating smart digital homes. Her publications include 4 papers in international journals and conferences and one national conference paper. She has 3 registered national patents and is now teaching Robocop robot designing for elementary and high school students at Alvand guidance school. She is a member of Exceptional Talented Students office of Razi University since 2008 and is a member of The Elite National Foundation of Iran.

Ali Reza Manashty is a M.S student in Artificial Intteligent Computer Engineering at Shahrood University of Thechnology, Shahrood, Iran. He got his B.S. degree in software computer engineering from Razi University ,Kermanshah , Iran, in 2010. He has been researching on mobile application design and smart environments especially smart digital houses since 2009. His publications include 4 papers in international journals and conferences and one national conference paper. He has earned several national and international awards regarding mobile applications developed by him or under his supervision and registered 4 national patents. He is a member of Exceptional Talented Students office of Razi University since 2008 and is a member of The Elite National Foundation of Iran. He was the teacher assistant of several under-graduate courses since 2008.